# Title: Boundary-Enhanced Segmentation of Pig Point Clouds in Commercial Housing Environments

**Zhankang Xu[a, d, 1], Fei Shi[a, c, 1], Xiangyu Qi[a], Zhaoyang Wang[a], Mengxin Guo[a], Yikai Fan[a], Simon X. Yang[b], Qifeng Li[*, a], Weihong Ma[*, a, b]**

[a] Information Technology Research Center, Beijing Academy of Agriculture and Forestry Sciences, Beijing, 100097, China

[b] Advanced Robotics and Intelligent Systems Laboratory, School of Engineering, University of Guelph, Guelph, ON N1G 2W1, Canada

[c] School of Science, China University of Geosciences (Beijing), Beijing 100083, China

[d] Bureau of Agriculture and Rural Affairs of Shangrao City, Jiangxi, China

[1] Zhankang Xu and Fei Shi contributed equally to this work

[*] Corresponding author.

E-mail address: nercita1017@163.com (Q.L.); mawh@nercita.org.cn (W.M.); xuzhankangs@163.com (Z.X.); shifei916@gmail.com (F.S.); qxy@cau.edu.cn (X.Q.); cauwzy@cau.edu.cn (Z.W.); guomengxin111@163.com (M.G.); fanyk@nercita.org.cn (Y.F.); syang@uoguelph.ca (Y.S.);

**Abstract:** In real pigsty environments, pig point clouds often come into close contact with background structures, resulting in blurred target boundaries, local adhesion, and background mis-segmentation. This reduces the accuracy of subsequent point cloud completion and body size measurement. To address these challenges, this study proposes a pig point cloud segmentation method based on boundary feature analysis. The proposed method adopts Octree Transformer as the backbone network and integrates local geometric details with global semantic context through octree convolution, self-attention encoding, and multi-scale feature fusion. Furthermore, soft-distance boundary pseudo-labels are generated to provide continuous boundary supervision, and a bidirectional cross-boundary semantic module is designed to enable explicit interaction between boundary and semantic features. Experiments conducted on a comprehensive dataset demonstrate that the proposed method significantly outperforms various state-of-the-art models in terms of segmentation accuracy, mean intersection over union, and boundary delineation. The results indicate that the method effectively alleviates boundary adhesion, providing reliable point cloud inputs for downstream precision livestock farming tasks.

**Keywords:** pig point cloud segmentation; boundary-aware learning; Octree Transformer; soft-distance boundary pseudo-label; precision livestock farming

## 1 Introduction

With the development of three-dimensional (3D) vision, deep learning, and intelligent perception technologies, point clouds—capable of directly representing the spatial geometric structure of objects—have been widely applied in 3D reconstruction, object recognition, scene understanding, and smart agriculture. Compared with two-dimensional (2D) images, point clouds not only record the 3D coordinates of target surfaces but also capture object contours, local surface details, and spatial topological relationships. Consequently, they are particularly suitable for acquiring animal phenotypic information and measuring body size parameters[1-3]. In pig farming scenarios, acquiring pig body point clouds using RGB-D cameras or multi-view depth cameras provides a critical data foundation for monitoring growth, assessing health, analyzing production performance, and performing non-contact body size measurements[4,5].

However, original point clouds captured in real pigsty environments typically include not only pig bodies but also background structures such as the floor, fences, railings, and data acquisition devices. As pigs are non-rigid living targets, postural changes—including head turning, walking, pausing, and approaching railings—frequently occur during data collection, often leading to local adhesion between pig body point clouds and background point clouds. Additionally, occlusion, sensor noise, uneven point cloud density, and background interference further compromise the geometric consistency of pig body boundaries, resulting in blurred class transitions, complex local structures, and unstable feature distributions at the boundaries. If pig body point clouds cannot be accurately separated, residual background points or missing edges of the pig body will continue to affect subsequent processes such as point cloud completion, keypoint detection, and body size measurement, thereby undermining the reliability of three-dimensional phenotypic measurements of pigs. Consequently, fine segmentation of pig point clouds under complex backgrounds is a critical pre-processing step for automated acquisition of three-dimensional phenotypes.

Early methods for animal point cloud segmentation primarily relied on strategies such as background modeling, Euclidean clustering, RANSAC plane removal, and manually defined geometric rules. These approaches can achieve satisfactory results when the target and background are clearly separated and the acquisition environment is relatively controlled[4,6,7]. However, such methods typically depend on fixed thresholds or hand-crafted features, limiting their adaptability to variations in pig postures, railing occlusions, point cloud density fluctuations, and boundary adhesion. In recent years, deep learning has facilitated the transition of point cloud segmentation from rule-driven approaches to data-driven feature learning. PointNet and PointNet++ laid the foundation for directly processing unordered point sets[8,9], while methods such as DGCNN, KPConv, Point Transformer, PointNeXt, and OctFormer further enhanced the modeling of local geometric relationships, long-range dependencies, and multi-scale structures in point clouds[10-14]. In the context of livestock and poultry point clouds, Lin et al.[15] employed improved DBSCAN and PointNet++ to extract pig point clouds, Wang et al.[16] combined 2D segmentation results with depth information to acquire pig body point clouds, and Chang et al.[17] enhanced segmentation of pig body and railing point clouds in complex pig farms by improving PointNet++ with SoftPool.

Although the aforementioned studies have advanced the automation of pig point cloud segmentation, mis-segmentation still predominantly occurs at contact boundaries between pig bodies and background objects such as the floor, fences, and railings in real breeding environments. Existing general point cloud segmentation networks typically emphasize overall semantic classification accuracy or enhance segmentation performance in main regions by improving local feature extraction capabilities. However, they often lack sufficient modeling of fine-grained geometric differences, category transition relationships, and boundary uncertainties at contact regions, which frequently leads to issues such as misclassifying background points as pig bodies, missing pig body edges, and local adhesion[17-19]. In particular, in regions where the pig's head contacts the railing, the legs contact the floor, or the side of the torso overlaps with a fence, it is difficult to obtain stable and well-defined boundaries relying solely on conventional local neighborhood aggregation or global semantic representations. Therefore, it is necessary to introduce additional supervision constraints and semantic-boundary interaction mechanisms for contact boundaries, based on an efficient point cloud feature extraction framework.

To address these issues, this paper proposes a pig point cloud segmentation method based on boundary feature analysis. The proposed method adopts Octree Transformer as the backbone network, organizes the original point cloud using an octree-based hierarchical structure, and incorporates multi-scale feature extraction and cross-layer fusion mechanisms to model both local geometric details and global semantic context in pig point clouds. Furthermore, soft-distance boundary pseudo-labels are generated to provide continuous and smooth boundary supervision signals for contact regions between the pig body and the background. A bidirectional cross-boundary semantic module is further designed to enable explicit interaction between boundary features and semantic features, thereby enhancing the model's discriminative capability for complex contact boundaries. The main contributions of this paper are summarized as follows:

(1) To address the problem of boundary adhesion between pig body and background point clouds in complex pigsty environments, a pig point cloud segmentation framework based on Octree Transformer was developed to jointly capture local geometric details and global semantic context.

(2) A soft-distance boundary pseudo-label is proposed to convert discrete boundary supervision into continuous boundary constraints, aiming to mitigate training instability caused by fuzzy category transitions near boundaries and variations in point cloud density.

(3) A bidirectional cross-boundary semantic module is designed to facilitate interaction and enhancement between semantic features and boundary features, thereby improving the model's fine-grained segmentation capability for contact regions between the pig body and background structures such as the floor, railings, and fences.

# 2 Materials and Methods

## 2.1 Data Acquisition

The point cloud data of live pigs used in this study were collected from Tieqilishi Pig Farm in Mianyang City, Sichuan Province, and a large-scale pig farm in Longyao County, Xingtai City, Hebei Province. The experimental subjects were Duroc × Landrace × Large White three-way crossbred pigs, which are widely used in large-scale pig farming and are highly representative. A total of 440 sets of point cloud data were acquired, with body weights ranging from 25 to 150 kg, covering the early, middle, and late fattening stages of pig growth. To minimize the influence of feeding state on pigs' external morphology, data collection was uniformly conducted before daily feeding, between 6:30 and 8:30.

The data acquisition zone was constructed within the main passageway of a standard, partitioned barn, configured with consistent artificial illumination, directional guide fences, infrared photoelectric sensors, and a radio frequency identification (RFID) reader. As the animal advanced into the detection zone, it interrupted the infrared barrier, thereby triggering real-time, synchronous point cloud acquisition. Concurrently, the RFID reader interrogated the electronic ear tag to automatically cross-reference individual identification (ID) with its corresponding spatial dataset—a practice increasingly pivotal for individual tracking, historical data association, and automated herd management in modern precision livestock farming[20]. Throughout the data collection campaign, the floors and adjacent fences served as fixed spatial reference points, and adverse environmental anomalies, such as non-uniform shading and specular reflections, were strictly controlled. Ultimately, the acquired datasets were restricted to 3D spatial coordinates (XYZ), whereas color channels (RGB) were intentionally omitted. This geometric-only representation successfully minimized computational overhead, optimized storage efficiency, and decoupled the downstream deep learning models from changing illumination conditions. The schematic layout of the acquisition hardware and the field data collection environment are illustrated in Figure. 1.

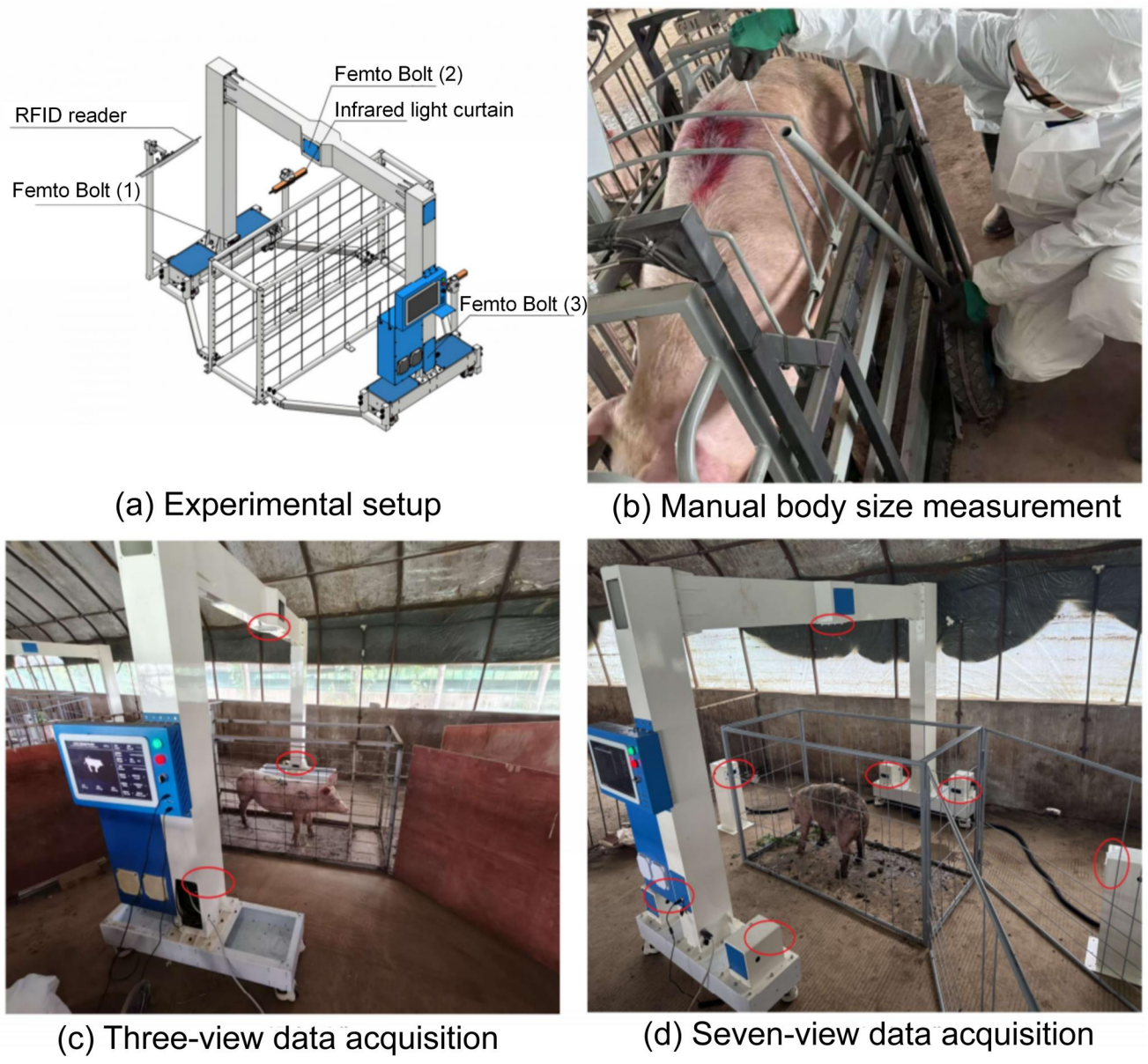

**Figure 1.** Data acquisition system and field setup for pig point cloud collection. The system integrates multi-view depth cameras, infrared triggering, directional guide fences, and RFID-based individual identification for synchronized point cloud acquisition in a commercial pig housing environment.

To balance practical acquisition efficiency and the integrity of point cloud structures, this study developed two types of synchronous point cloud acquisition systems with three-view and seven-view configurations. The three-view system consists of three Orbbec Femto Bolt depth cameras, which are arranged above and on the left and right sides of the pig body, respectively, forming a spatial observation structure from the upper and bilateral directions. The cameras operate in WFOV 2×2 binned mode, with a field of view of 120° × 120°, an effective measurement range of 0.25–2.88 m, and an acquisition frame rate of 30 fps. The distance from the lens of the top camera to the ground is approximately 1.6 m, the distance between the lenses of the two side cameras is approximately 2.0 m, and the width of the acquisition channel is approximately 0.5 m, allowing pigs to pass through in a single file. The three-view system can acquire point clouds of the dorsal and lateral regions of the pig body; however, local missing areas may still occur in the abdomen, the inner sides of the limbs, and regions occluded by railings. An example of a three-view pig point cloud is shown in Figure 2.

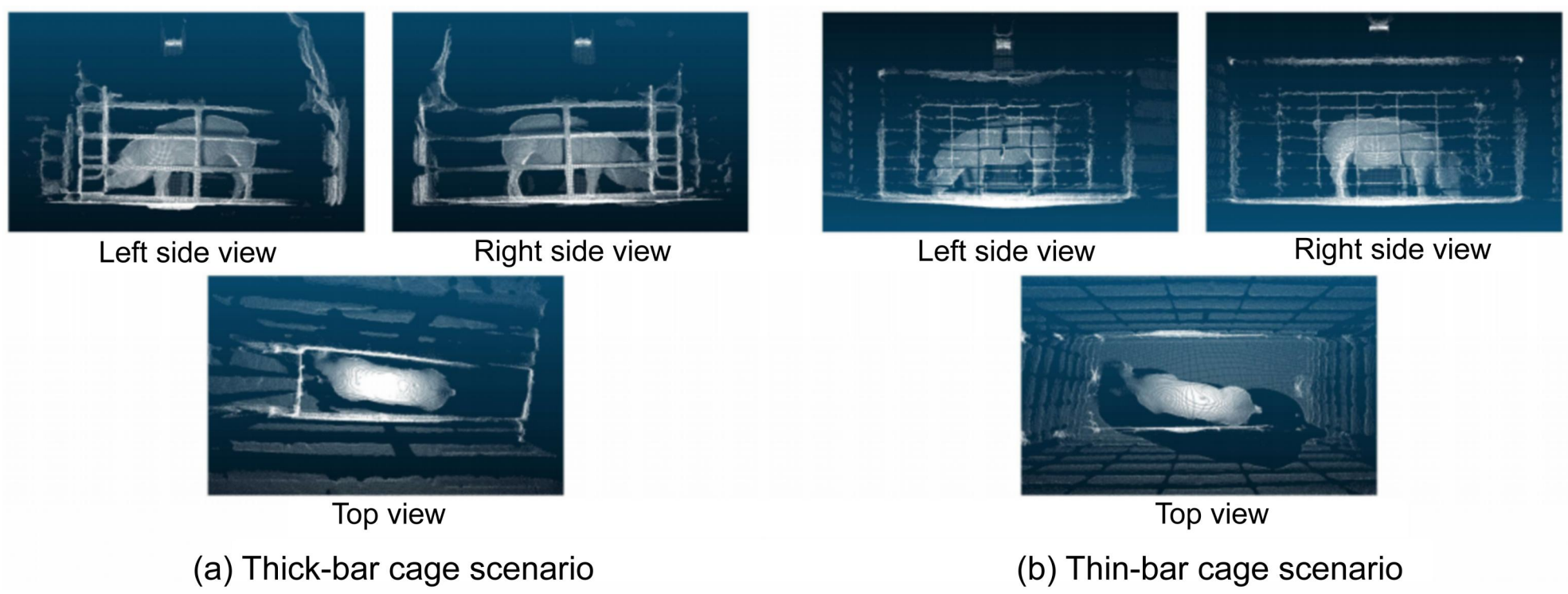

**Figure 2.** Representative three-view pig point cloud. The three-view configuration captures dorsal and bilateral surface information of the pig body, while local missing regions may remain in occluded areas such as the abdomen, inner limbs, and railing-contact regions.

To further mitigate the impact of single-view occlusion on the integrity of pig point clouds, this study developed a seven-view synchronous point cloud acquisition system. The system comprises seven depth cameras arranged according to an optimized spatial layout, which synchronously capture pig point clouds from the top, left front, right front, left rear, right rear, front, and rear directions, forming a multi-directional redundant observation structure. The camera models, acquisition parameters, and triggering methods of the seven-view system are consistent with those of the three-view system, and hardware synchronization ensures that multi-camera data are acquired within the same time window, thereby minimizing the effects of slight pig movements on point cloud fusion. Compared with three-view data, seven-view point clouds exhibit higher coverage integrity and spatial continuity in typically occluded regions such as the abdomen, inner sides of the limbs, head, and buttocks, enabling a more comprehensive representation of the overall geometric shape of the pig body. During data acquisition, each pig passes through the passage 2–3 times to increase sample size and pose diversity. An example of a seven-view pig point cloud is shown in Figure 3.

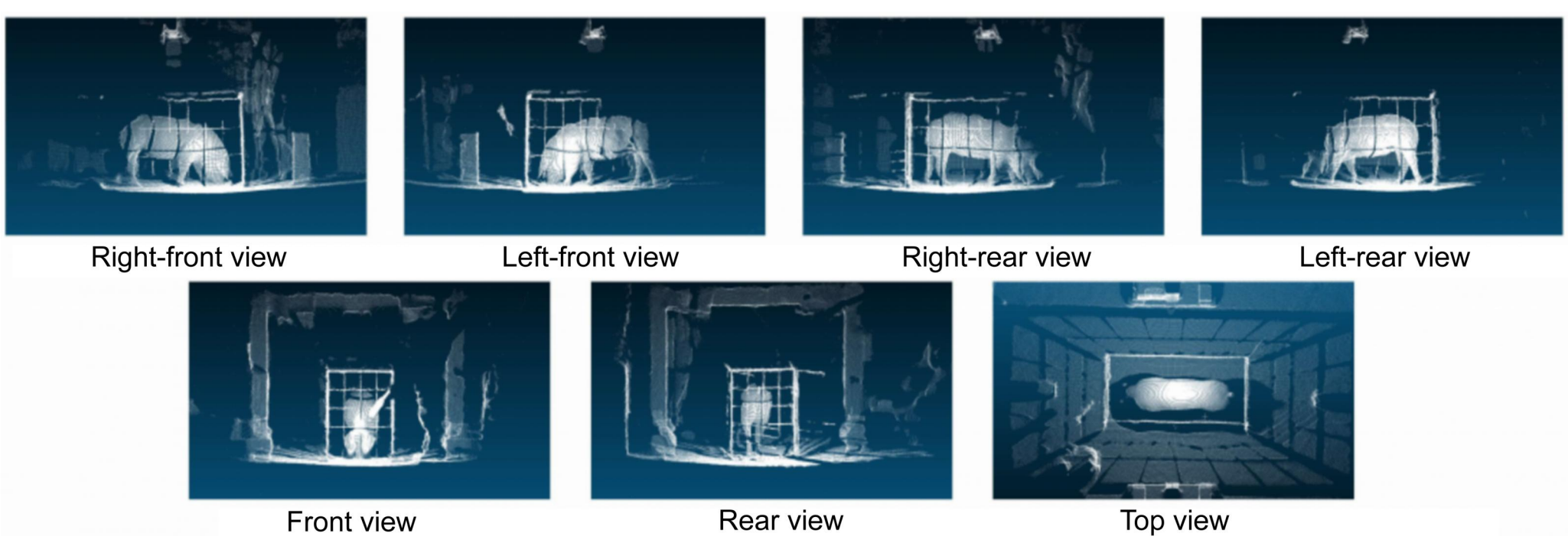

**Figure 3.** Representative seven-view pig point cloud. The seven-view configuration provides more complete spatial coverage of the pig body by capturing point clouds from the top, front, rear, and bilateral oblique directions.

## 2.2 Point Cloud Preprocessing and Annotation

To fuse multi-view point clouds within a unified coordinate system, this study adopted Zhang's checkerboard calibration method[21] to geometrically calibrate multiple synchronized cameras. During calibration, a 60 cm × 80 cm black-and-white checkerboard was used, and multiple sets of images with different poses were collected under unobstructed conditions. First, the focal length, principal point coordinates, and distortion parameters of each camera were estimated. Then, the extrinsic parameter relationships between cameras were obtained through pairwise chain calibration. The coordinate system of the top camera was used as the global reference coordinate system. Point clouds captured from different viewpoints were transformed into the unified world coordinate system, with the spatial unit standardized to centimeters (cm). In this coordinate system, the Z-axis is perpendicular to the ground, the X-axis follows the direction of the acquisition channel, and the Y-axis is parallel to the ground and perpendicular to the channel direction. Through the above calibration and coordinate transformation procedures, multi-view fused point clouds with consistent spatial positions were obtained, providing standardized inputs for subsequent segmentation annotation and model training.

After multi-view point cloud fusion, spatial range cropping and quality screening were performed on all samples. First, a region of interest (ROI) was defined according to the structural dimensions of the acquisition channel, and redundant background points clearly outside the pigs' activity range were removed, including points below the ground, points at the top of the channel, and discrete background regions far from the main body of the pigs. Subsequently, manual quality inspection was conducted on the point cloud samples, and distorted data caused by severely bent postures, jumping, motion blur, or sensor abnormalities were excluded. After screening, a total of 3,392 point cloud samples were obtained from the three-view and seven-view data, among which 1,042 were complete seven-view point clouds. All point clouds retained the corresponding pig IDs to ensure that data from the same pig under different views and at different acquisition times could be tracked and managed.

The point cloud segmentation dataset was manually annotated using CloudCompare software. Since the original point clouds contain both pig bodies and environmental structures, including the ground, fences, railings, and data acquisition devices, the pig point cloud segmentation task was defined as a binary classification problem: the pig body region was labeled as “1” and the environmental background as “0.” Given the boundary ambiguity in areas where the pig body contacts the railings or where the limbs meet the ground, manual annotation results may be influenced by subjective judgment. To minimize errors arising from inter-annotator differences, all samples were annotated by the same individual according to a standardized protocol.

For dataset partitioning, the training, validation, and test sets were divided using individual pigs as the basic unit, rather than by randomly splitting individual point cloud samples. This strategy was adopted to avoid data leakage caused by samples from the same pig, captured from different viewpoints or at different acquisition times, appearing simultaneously in both the training and test sets. The entire dataset was divided at a ratio of 8:1:1, with the training, validation, and test sets containing 352, 44, and 44 pigs, respectively. During partitioning, the balanced distribution of different body weight ranges was considered to improve the representativeness of the training and test samples. The detailed partitioning results are presented in Table 1.

**Table 1.** Dataset split for pig point cloud segmentation.

| Dataset | Number of pigs | Number of point cloud samples | Number of complete seven-view point cloud samples |
|---|---|---|---|
| Training set | 352 | 2714 | 834 |
| Validation set | 44 | 339 | 104 |
| Test set | 44 | 339 | 104 |
| Total | 440 | 3392 | 1042 |

**Note:** The dataset was split at the individual-pig level at a ratio of 8:1:1. “Number of complete seven-view point cloud samples” refers to the samples acquired using the seven-view system and retained after quality screening in each dataset.

## 2.3 Boundary-Aware Pig Point Cloud Segmentation

### 2.3.1 Overview of the Proposed Method

In real pigsty environments, pig point clouds often come into contact with background structures such as the floor, fences, and railings. Boundary regions are characterized by uneven point cloud density, complex geometric structures, and blurred semantic category transitions. When relying solely on conventional point cloud semantic segmentation networks, issues such as boundary adhesion, local segmentation omissions, or misclassification of background points as part of the pig body frequently arise at contact regions between the pig body and the environment. To address these challenges, this study proposes a pig point cloud segmentation method based on boundary feature analysis.

The proposed method adopts Octree Transformer[14] as the backbone network. First, the input point cloud is organized into an octree-based hierarchical structure, and multi-scale point cloud features are extracted through octree convolution and self-attention encoding modules. Subsequently, coarse-scale global semantic information and fine-scale local geometric information are integrated using a cross-layer feature fusion mechanism. On this basis, soft-distance boundary pseudo-labels are generated to provide continuous boundary supervision for contact regions between the pig body and the background. Finally, a bidirectional cross-boundary semantic module is employed to enable explicit interaction between semantic features and boundary features, and a joint loss function is adopted to simultaneously optimize semantic segmentation and boundary prediction. The overall network architecture is shown in Figure 4.

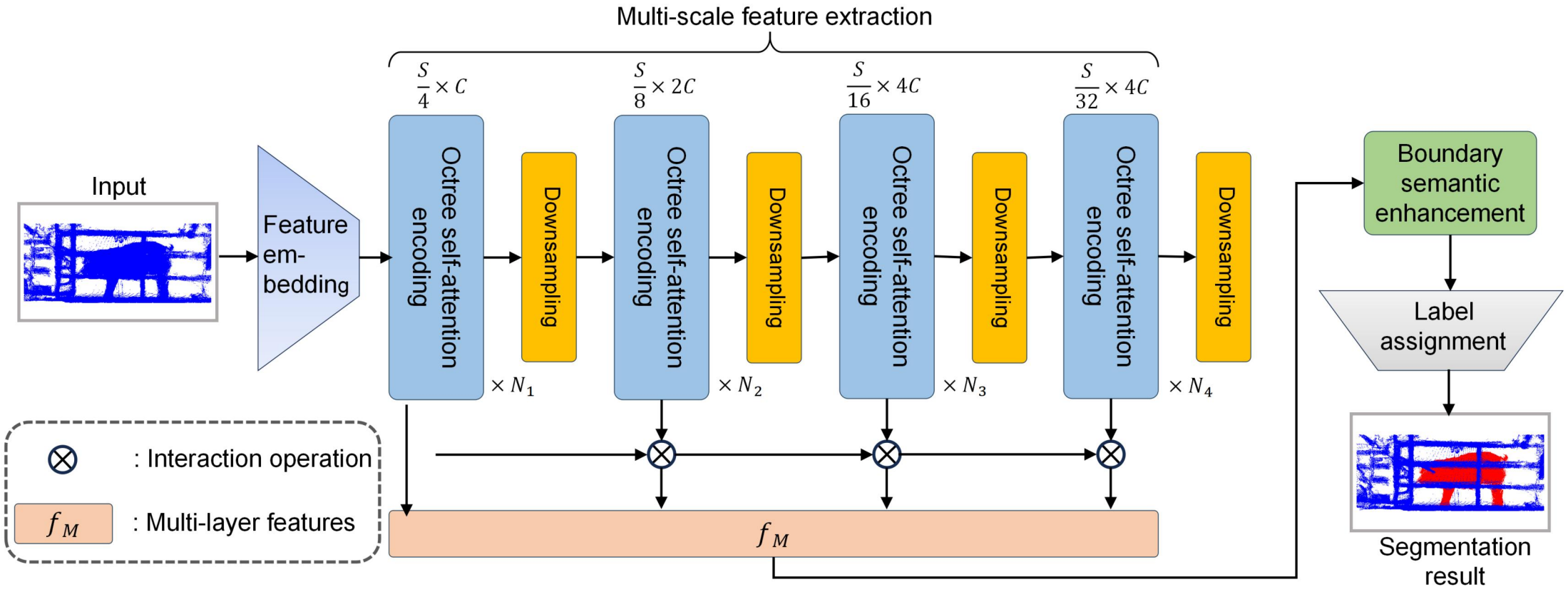

**Figure 4.** Overall architecture of the proposed boundary-aware pig point cloud segmentation network. The network uses Octree Transformer as the backbone and integrates multi-scale feature fusion, soft-distance boundary supervision, and bidirectional semantic-boundary feature interaction for pig point cloud segmentation.

2.3.2 Multi-Scale Point Cloud Feature Extraction

Point cloud data are characterized by unorderedness, sparsity, and irregular spatial distribution. Directly applying global self-attention to all points would incur substantial computational overhead. To improve the processing efficiency of large-scale pig point clouds, this study adopts the octree encoding strategy of OctFormer. Specifically, the original point cloud is first organized into an octree-based hierarchical structure, allowing adjacent non-empty nodes in three-dimensional space to participate in feature computation in a locally continuous manner.

In the feature embedding stage, the network employs five octree convolution modules to project the input point cloud features into a high-dimensional feature space. Each module consists of an octree convolution layer[22], batch normalization, and a ReLU activation function. The convolution kernel sizes are sequentially set to {3, 2, 3, 2, 3}, with corresponding strides of {1, 2, 1, 2, 1}. Among these modules, convolution layers with a stride of 2 are used for spatial downsampling. After five octree convolution layers, the point cloud features undergo an overall four-fold spatial downsampling, thereby reducing computational cost while enhancing local geometric representation capability.

In the octree self-attention encoding stage, the network is organized into multiple stages, each containing several octree self-attention encoding modules and downsampling modules. Each encoding module primarily comprises octree attention, a multi-layer perceptron (MLP), and layer normalization. To facilitate windowed attention computation, the Morton encoding strategy[23] is adopted to map adjacent non-empty octree nodes in three-dimensional space into a one-dimensional, locally continuous sequence. The sequence is then partitioned into fixed-length windows, within which self-attention is computed. To enhance cross-window information interaction, the network alternates dilation factors of $D = 1$ and $D = 4$ for window partitioning in consecutive self-attention layers. When $D = 1$, the model focuses primarily on detailed structures within local neighborhoods; when $D > 1$, the dilated windows span previously adjacent node blocks, thereby expanding the context receptive field. By alternately computing local and dilated windows, the model balances local boundary detail with large-scale semantic context while maintaining approximately linear computational complexity. The octree self-attention encoding module is illustrated in Figure 5.

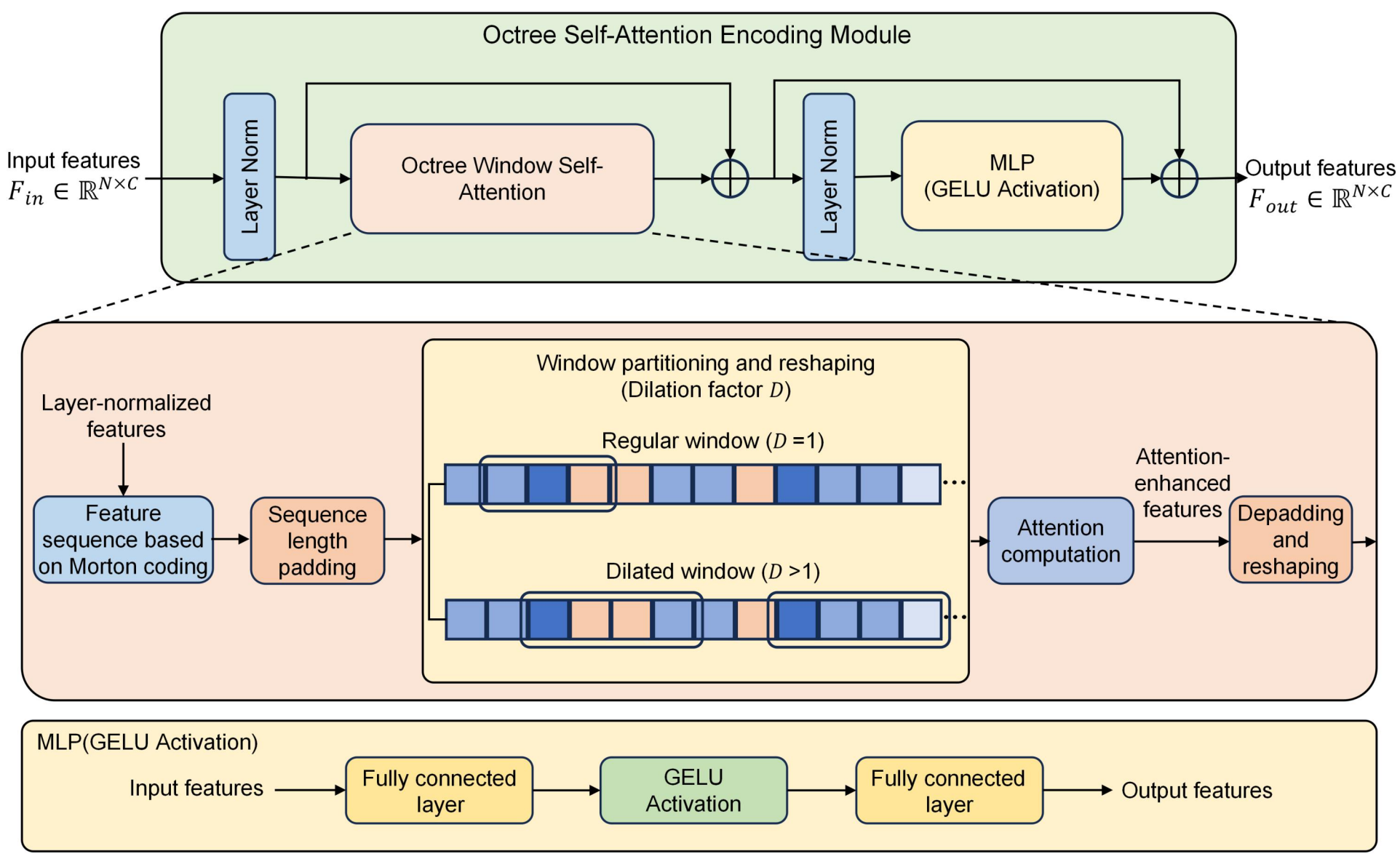

**Figure 5.** Architecture of the octree self-attention encoding module. Non-empty octree nodes are encoded using Morton ordering and processed through local and dilated window self-attention to capture both local geometric details and broader contextual information.

Considering that different octree levels correspond to different semantic scales, this study further implements multi-scale feature fusion. In general, higher-level nodes capture stronger global structural information, while lower-level nodes retain richer local boundary details. Inspired by the multi-resolution feature interaction approach of HRNet[24], cross-layer fusion is performed on features from the last four octree levels. Following the settings of OctFormer, the maximum octree depth is 11, and therefore, fusion is applied from the 8th to the 11th levels. The computation of the fused feature $f_M$ is defined as follows:

$$f_M = up(f_8) + \sum_{i=8}^{11} up\left(C^{3\times3}\left(up(f_i) \otimes C^{1\times1}(f_{i+1})\right)\right) \tag{2.1}$$

In the formula, $f_i$ represents the features of the *i-th* octree layer; $\otimes$ denotes the feature interaction operation; $C^{1\times1}$ and $C^{3\times3}$ represent 1×1 and 3×3 convolution layers, respectively; and *up(·)* denotes the upsampling operation. Through cross-layer fusion, a point cloud representation capturing both global semantic information and fine-grained geometric structures can be obtained.

### 2.3.3 Soft-Distance Boundary Pseudo-Labeling

In pig point cloud segmentation, class boundaries between pig bodies and background structures, such as the floor, fences, and railings, are often not strictly discrete but instead exhibit spatial transition regions. Traditional hard-boundary pseudo-labels typically classify points simply as boundary or non-boundary points according to whether heterogeneous points are present in their local neighborhoods. Although this approach is straightforward to implement, it has two main limitations. First, boundary thickness is highly sensitive to point cloud sampling density. Second, binary labels can only distinguish boundary points from non-boundary points and cannot represent the continuous distance relationship between points and the true boundary, which hinders the model from learning the gradual structural variations near boundaries.

To address the aforementioned issues, this study introduces the Soft Distance Boundary Pseudo-Label (SDBPL). This approach generates a continuous boundary probability based on the distance from a point to its nearest out-of-class point, providing a smoother supervision signal for boundary regions. For any point *i*, let its spatial coordinates be $C_i$ and its semantic label be $S_i$. Within the neighborhood $N_i(r)$ of radius *r*, the distance $d_i$ from this point to its nearest out-of-class point is computed as follows:

$$d_i = \min_{j \in N_i(r), S_j \neq S_i} \left\| C_i - C_j \right\|_2 \tag{2.2}$$

In the formula, $\|\cdot\|_2$ denotes the Euclidean distance. If no out-of-class points exist within the neighborhood, let $d_i = +\infty$. Based on this distance, the soft-boundary pseudo-label $B_i^g \in [0，1]$ is defined as follows:

$$B_i^g = \begin{cases} \exp\left(-\dfrac{d_i^2}{2\sigma^2}\right), & d_i \leq r \\ 0, & else \end{cases} \tag{2.3}$$

In the formula, σ controls the width of the boundary transition zone. As $d_i$ approaches 0, the closer point $i$ is to the true boundary, the closer its boundary probability is to 1. As $d_i$ increases, the boundary probability gradually decays toward 0. Compared with binary hard-boundary labels, soft distance boundary pseudo-labels form a continuous and differentiable constraint near boundaries, which not only preserves precise boundary position information but also provides effective supervision signals for the boundary transition regions, thereby enhancing the model's ability to represent complex boundary contours, such as contact areas involving the abdomen, limbs, and railings. A schematic diagram of the soft distance boundary pseudo-label is illustrated in Figure 6.

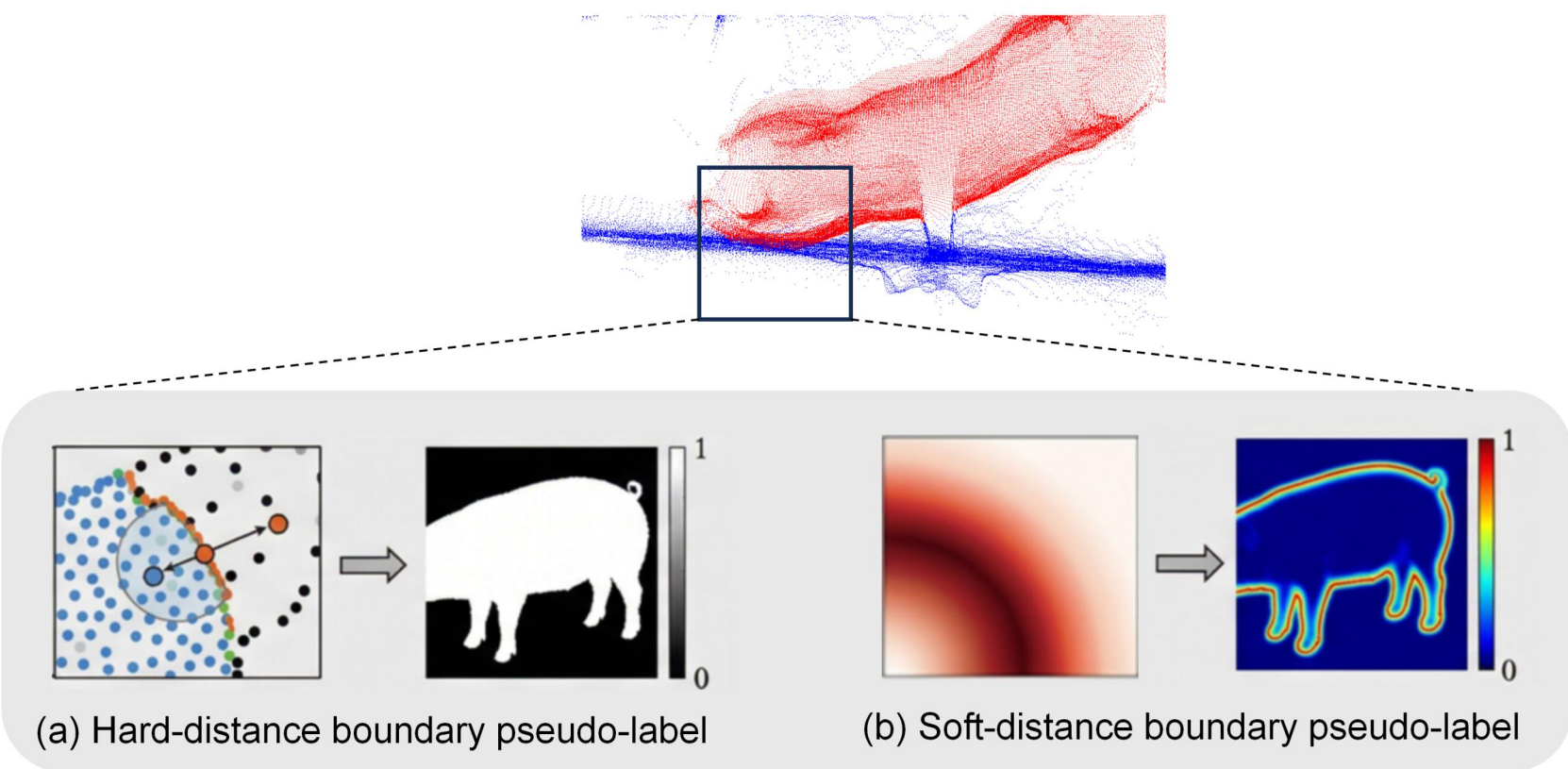

**Figure 6.** Illustration of the soft-distance boundary pseudo-labeling strategy. Boundary supervision is generated according to the distance from each point to its nearest out-of-class point, producing a continuous boundary probability that decays from the contact boundary to non-boundary regions.

2.3.4 Bidirectional Cross Boundary–Semantic Enhancement Module

Relying solely on multi-scale feature fusion remains insufficient to fully resolve the semantic adhesion problem in boundary regions. Existing boundary enhancement methods typically leverage boundary features to guide semantic attention computation[19]; however, boundary structural information does not fully participate in updating semantic features. When boundary prediction results are unstable or multi-scale ambiguities exist, unidirectional information flow is insufficient to effectively mitigate category confusion in contact regions between the pig body and the background.

To enhance the interaction between semantic features and boundary features, this study designs a bidirectional cross-boundary semantic module. This module first separates the multi-scale fused features into semantic and boundary features, and then facilitates explicit interaction between them through bidirectional cross-attention. Unlike unidirectional boundary guidance methods, boundary features in this module not only act as queries in attention computation but also participate as keys and values in updating semantic features, thereby improving the transfer of boundary structural information into the semantic space. The bidirectional cross-boundary semantic module is illustrated in Figure 7.

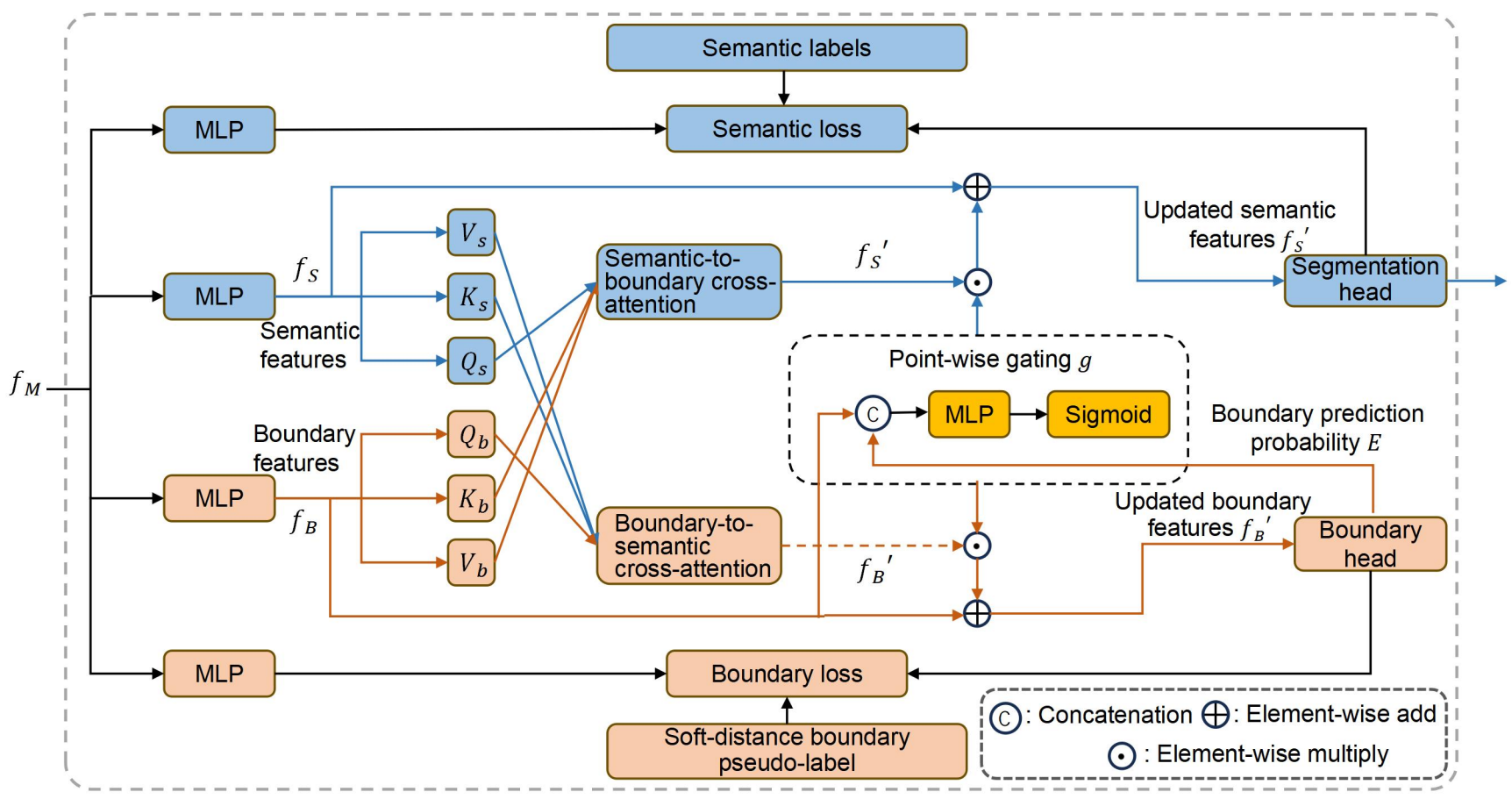

**Figure 7.** Architecture of the bidirectional cross boundary–semantic enhancement module. Semantic and boundary features are updated through bidirectional cross-attention and point-level gating, enabling adaptive information exchange between the semantic branch and the boundary branch.

Specifically, given an input feature $f_M \in R^{N\times C}$, it is first decomposed into a semantic feature $f_S$ and a boundary feature $f_B$ through two independent mapping functions.

$$f_S = \phi_S(f_M),\ f_B = \phi_b(f_M) \tag{2.4}$$

In the formula, both $\phi_b(\cdot)$ and $\phi_S(\cdot)$ consist of multi-layer perceptrons. Subsequently, the queries, keys, and values required for multi-head attention are constructed as follows:

$$\begin{aligned} Q_S &= f_S W_q^S, & K_S &= f_S W_k^S, & V_S &= f_S W_v^S, \\ Q_b &= f_B W_q^b, & K_b &= f_B W_k^b, & V_b &= f_B W_v^b. \end{aligned} \tag{2.5}$$

On this basis, bidirectional cross-attention is applied to mutually enhance both semantic and boundary features.

$$\Delta f_S = \mathrm{Softmax}\left(\frac{Q_S K_b^T}{\sqrt{d}}\right) V_b \tag{2.6}$$

$$\Delta f_B = \mathrm{Softmax}\left(\frac{Q_b K_S^T}{\sqrt{d}}\right) V_S \tag{2.7}$$

In the formula, $d$ denotes the feature dimension and is used to scale the inner product, thereby stabilizing the training process. To suppress interference from noisy boundary predictions on the semantic representation of non-boundary regions, a point-level gating vector $g$ is further introduced:

$$g = \sigma(MLP([f_B, E])) \tag{2.8}$$

Among them, $E$ denotes the boundary prediction probability, $[\cdot,\cdot]$ represents feature concatenation, and $\sigma(\cdot)$ is the Sigmoid function. Finally, the semantic and boundary features are updated as follows:

$$f_S' = f_S + g \odot \Delta f_S \tag{2.9}$$

$$f_B' = f_B + (1-g) \odot \Delta f_B \tag{2.10}$$

In the formula, $\odot$ denotes element-wise multiplication. This gating mechanism enables the model to adaptively regulate information flow according to point-wise boundary confidence. Specifically, when a point has a high boundary probability, the semantic branch more effectively incorporates boundary structural information; when a point is far from the boundary, the gating mechanism suppresses the interference of boundary information with semantic representation, while the boundary branch obtains a more stable structural prior from the semantic context. Through this mechanism, the model achieves stronger boundary discrimination capability in contact regions between the pig body and background structures such as the floor and railings.

2.3.5 Joint Loss Function

To simultaneously optimize pig semantic segmentation and boundary prediction, this study jointly employs cross-entropy (CE) loss[25], binary cross-entropy (BCE) loss, and Dice loss[26] to construct a joint loss function during training, thereby enabling supervised learning for both the semantic prediction branch and the boundary prediction branch. Given a predicted distribution $X$ containing $N$ points and the corresponding ground-truth label $Y$, Dice loss is defined as follows:

$$\mathrm{Dice}(X,Y)=1-\frac{2\sum_{i=1}^{N}X_i^T Y_i}{\sum_{i=1}^{N}\left(X_i^T X_i+Y_i^T Y_i\right)} \tag{2.11}$$

The semantic prediction score is denoted as $P \in [0, 1]^{N\times M}$, where $N$ represents the number of points and $M$ represents the number of semantic classes. To simultaneously account for point-wise classification accuracy and the consistency of overlapping predicted regions, cross-entropy loss and Dice loss are combined to construct the semantic segmentation loss $L_{sem}$.

$$L_{sem}=\frac{1}{N}\sum_{i=1}^{N}CE\left(P_i,P_i^g\right)+Dice\left(P_i,P_i^g\right) \tag{2.12}$$

In the formula, $P_i$ denotes the semantic prediction probability of the *i-th* point, and $P_i^g$ denotes its corresponding ground-truth semantic label. For the boundary prediction task, the network outputs the probability of each point being a boundary point, and the boundary score is denoted as $E \in [0, 1]^{N\times 1}$, where $E_i$ represents the boundary prediction probability of the *i-th* point. The boundary loss $L_{bou}$ is composed of binary cross-entropy loss and Dice loss:

$$L_{bou}=\frac{1}{N}\sum_{i=1}^{N}BCE\left(E_i,E_i^g\right)+Dice\left(E_i,E_i^g\right) \tag{2.13}$$

In the formula, $E_i^g$ denotes the soft distance boundary pseudo-label corresponding to the *i-th* point. Finally, the overall loss function is defined as follows:

$$L=L_{sem}+\lambda L_{bou} \tag{2.14}$$

In the formula, λ denotes the weight coefficient, which balances the contributions of the semantic segmentation loss and the boundary loss in the overall optimization process. Through joint optimization, the network enhances its ability to perceive complex boundary regions while simultaneously learning the overall semantic structure of the pig body, thereby improving the precision of pig point cloud segmentation.

## 2.4 Experimental Setup and Evaluation Metrics

The experiment utilized the previously described pig point cloud segmentation dataset for model training, validation, and testing. Specifically, the training set was employed for network parameter learning, the validation set for model selection and parameter tuning, and the test set solely for final segmentation performance evaluation. The experimental hardware platform consisted of an Intel(R) Xeon(R) Platinum 8375C CPU, an NVIDIA RTX A6000 GPU, and 128 GB of system memory. The operating system was Ubuntu 20.04.6 LTS, and the deep learning framework was PyTorch 1.13.0. Model training was conducted using the AdamW optimizer, with an initial learning rate of 0.001 and a weight decay of 0.05 to mitigate overfitting and enhance model generalization. The experimental operating environment is summarized in Table 2.

**Table 2.** Experimental environment configuration.

| Development environment | Configuration |
|---|---|
| Processor | Intel(R) Xeon(R) Platinum 8375C CPU @ 2.90 GHz |
| GPU | NVIDIA RTX A6000 with 48 GB memory |
| RAM | 128 GB |
| Operating system | Ubuntu 20.04.6 LTS |
| Development framework | PyTorch 1.13.0 |

To comprehensively evaluate the model's performance in pig point cloud segmentation, this study adopts accuracy (*Acc*), mean intersection over union (*mIoU*), and boundary intersection over union (*B-IoU*[27]) as evaluation metrics. Specifically, Acc measures the overall classification correctness of all points, *mIoU* evaluates the average degree of overlap between the predicted regions and ground-truth labels across different classes, and *B-IoU* further quantifies the boundary segmentation quality in contact regions between the pig body and background structures such as the floor, fences, and railings. Accuracy (*Acc*) is defined as follows:

$$Acc = \frac{TP + TN}{TP + TN + FP + FN} \tag{2.15}$$

In the formula, *TP* denotes the number of points correctly classified as pig body points, *TN* denotes the number of points correctly classified as background points, *FP* denotes the number of background points misclassified as pig body points, and *FN* denotes the number of pig body points misclassified as background points. Intersection over Union (*IoU*) measures the degree of overlap between predicted results and ground-truth labels for a given category. For category *C*, its *IoU* is defined as follows:

$$IoU_c = \frac{|\mathrm{Prediction}_c \cap \mathrm{GroundTruth}_c|}{|\mathrm{Prediction}_c \cup \mathrm{GroundTruth}_c|} \tag{2.16}$$

In this binary classification task, the number of categories ($C$ = 2) corresponds to the pig body and background categories. Therefore, the mean Intersection over Union (*mIoU*) is defined as follows:

$$mIoU = \frac{1}{C}\sum_{c=1}^{C} IoU_c \tag{2.17}$$

Since this study focuses on boundary segmentation in contact regions between pigs and background structures such as railings and floors within complex pigsty environments, accuracy (*Acc*) and mean Intersection over Union (*mIoU*) alone may not fully capture fine-grained missegmentation at boundaries. Therefore, *B-IoU* is additionally adopted to evaluate the model's boundary recognition capability. First, the sets of ground-truth boundary points $B_l$ and predicted boundary points $B_p$ are determined based on the neighborhood relationships of points: a point is considered a boundary point if its neighborhood contains points of different classes. Following previous studies, the neighborhood radius is set to $r = 0.1m$[11,28]. *B-IoU* is defined as follows:

$$B\text{-}IoU = \frac{|B_l \cap B_p|}{|B_l \cup B_p|} \tag{2.18}$$

In the formula, $B_p$ denotes the set of boundary points obtained from model predictions, and $B_l$ denotes the set of boundary points corresponding to the ground-truth labels. A higher *B-IoU* indicates that the model can more accurately identify boundaries in contact regions between the pig body and background structures such as railings and the floor, thereby more effectively reducing boundary adhesion, background residues, and missed segmentation of pig body edges.

# 3 Results and Analysis

## 3.1 Experimental Design and Comparative Methods

To evaluate the performance of the proposed method for pig point cloud segmentation in complex pigsty environments, this study conducts comparative experiments using the same data splits and standardized evaluation metrics. The training set is employed for network parameter learning, the validation set for model selection, and the test set for final performance assessment. The evaluation metrics include accuracy (*Acc*), mean Intersection over Union (*mIoU*), and boundary Intersection over Union (*B-IoU*). Specifically, Acc reflects the overall classification accuracy of all points, *mIoU* measures the average overlap between predicted regions and ground-truth labels, and *B-IoU* evaluates the boundary segmentation quality in contact regions between the pig body and background structures such as the floor, fences, and railings. Because this method primarily addresses issues such as boundary adhesion, local missed segmentation, and misclassification of background points as pig body points, *B-IoU* provides a more direct assessment of the model's segmentation capability at complex contact boundaries.

To ensure a fair experimental comparison, this study selected five representative point cloud semantic segmentation models as comparative methods: PointNet++, OctFormer, OA-CNNs, PointStack, and Point Transformer V3 (PTv3). Among them, PointNet++ is a classic hierarchical point cloud network that performs semantic segmentation on unordered point sets through sampling, grouping, and local feature aggregation, serving as a fundamental baseline for point cloud segmentation[9]. OctFormer organizes point clouds using an octree structure and performs Transformer-based attention computation within local windows, thereby enhancing point cloud feature representation while reducing computational complexity[14]. OA-CNNs improve the modeling of local structures in sparse point clouds through adaptive receptive fields and adaptive relationship convolution[29]. PointStack aggregates point features from multiple layers at different resolutions while preserving both high-level semantic information and high-resolution details[30]. PTv3 expands the receptive field by serializing point clouds and simplifying the Transformer structure, thereby improving the efficiency of point cloud feature computation[31].

The comparison models encompass diverse technical approaches, including hierarchical point networks, octree transformers, convolutional point cloud networks, multi-resolution feature aggregation networks, and serialized transformers, allowing evaluation of the proposed method's effectiveness in pig point cloud segmentation from multiple perspectives. To minimize the influence of non-model factors on experimental results, all comparison models are trained and tested using the same training, validation, and test set splits, with model-specific hyperparameters adjusted according to the recommended settings of the original methods.

## 3.2 Comparative Experiments on Point Cloud Segmentation

The quantitative evaluation results of each model on the pig point cloud segmentation test set are presented in Table 3. The proposed method achieves the best performance across all three metrics, with *Acc*, *mIoU*, and *B-IoU* reaching 98.65%, 97.36%, and 88.77%, respectively. Compared with the second-best model, OctFormer, the proposed method improves *Acc* by 0.55 percentage points, *mIoU* by 0.35 percentage points, and *B-IoU* by 3.37 percentage points. Since *Acc* and *mIoU* are already relatively similar among high-performing models, the room for improvement in overall classification accuracy is limited. In contrast, *B-IoU* more directly reflects the boundary prediction quality in contact regions between the pig body and background structures such as the floor, fences, and railings. Therefore, the improvement in *B-IoU* indicates that the main advantage of the proposed method lies not merely in improving the classification accuracy of large pig body regions but in enhancing fine-grained discrimination at complex contact boundaries.

**Table 3.** Comparative experimental results for pig point cloud segmentation.

| Model | Acc (%)↑ | mIoU (%)↑ | B-IoU (%)↑ |
|---|---|---|---|
| PointNet++ | 95.10 | 93.89 | 70.20 |
| OctFormer | 98.10 | 97.01 | 85.40 |
| OA-CNNs | 97.98 | 96.10 | 80.89 |
| PointStack | 96.45 | 95.02 | 76.09 |
| PTv3 | 98.02 | 96.89 | 83.10 |
| Proposed method | **98.65** | **97.36** | **88.77** |

**Note:** Acc, accuracy; mIoU, mean Intersection over Union; B-IoU, boundary Intersection over Union. ↑ indicates that a higher value represents better performance.

Among the comparison models, PointNet++ achieves *Acc* and *mIoU* values of 95.10% and 93.89%, respectively, whereas its *B-IoU* is only 70.20%. This indicates that PointNet++ can perform basic binary segmentation of the pig body and background but has limited capability in recognizing complex boundary regions. This result is closely related to its network architecture: PointNet++ primarily relies on hierarchical sampling, local neighborhood grouping, and feature aggregation for point cloud representation, which limits its ability to model fine structural variations and category transition relationships at contact regions between the pig body, railings, and the floor. Consequently, it is more prone to misclassifying background points as pig body points or missing pig body edges in boundary adhesion regions.

OctFormer organizes point clouds using an octree structure and performs self-attention computations within local windows. Its overall segmentation performance surpasses that of PointNet++, with *Acc* and *mIoU* reaching 98.10% and 97.01%, respectively. However, OctFormer's *B-IoU* is 85.40%, still lower than that achieved by the proposed method. This indicates that although a general octree Transformer backbone can enhance overall semantic feature representation, it remains challenging to fully resolve boundary confusion in contact regions between the pig body and complex backgrounds without explicit boundary supervision and a semantic-boundary interaction mechanism.

OA-CNNs, PointStack, and PTv3 enhance point cloud feature representation from the perspectives of adaptive convolution, multi-resolution feature aggregation, and serialized Transformer, respectively. Among these models, OA-CNNs achieve a *B-IoU* of 80.89%, indicating that adaptive receptive fields and local geometric modeling contribute to boundary recognition. PTv3 attains *Acc* and *mIoU* values of 98.02% and 96.89%, respectively, with a *B-IoU* of 83.10%, suggesting that serialized Transformers improve context modeling but still lag behind the proposed method in boundary segmentation quality. PointStack achieves a *B-IoU* of 76.09%, highlighting that relying solely on multi-resolution feature stacking is insufficient to fully address complex boundary adhesion at contact regions between the pig body and railings or the floor.

Overall, the proposed method introduces soft distance boundary pseudo-labels and a bidirectional cross-boundary semantic module on top of OctFormer's multi-scale feature representation, enabling the model to simultaneously capture local geometric information, global semantic context, and continuous boundary supervision. The soft distance boundary pseudo-labels alleviate training instability caused by abrupt changes of hard labels near boundaries, while the bidirectional cross-boundary semantic module facilitates information interaction between semantic and boundary features, thereby enhancing the model's discriminative capability for complex contact boundaries. Consequently, the proposed method achieves a notable advantage in the *B-IoU* metric, indicating its suitability for segmenting pig point clouds in contact with background structures such as railings and the floor within real pigsty environments. The performance comparison of different models is illustrated in Figure 8.

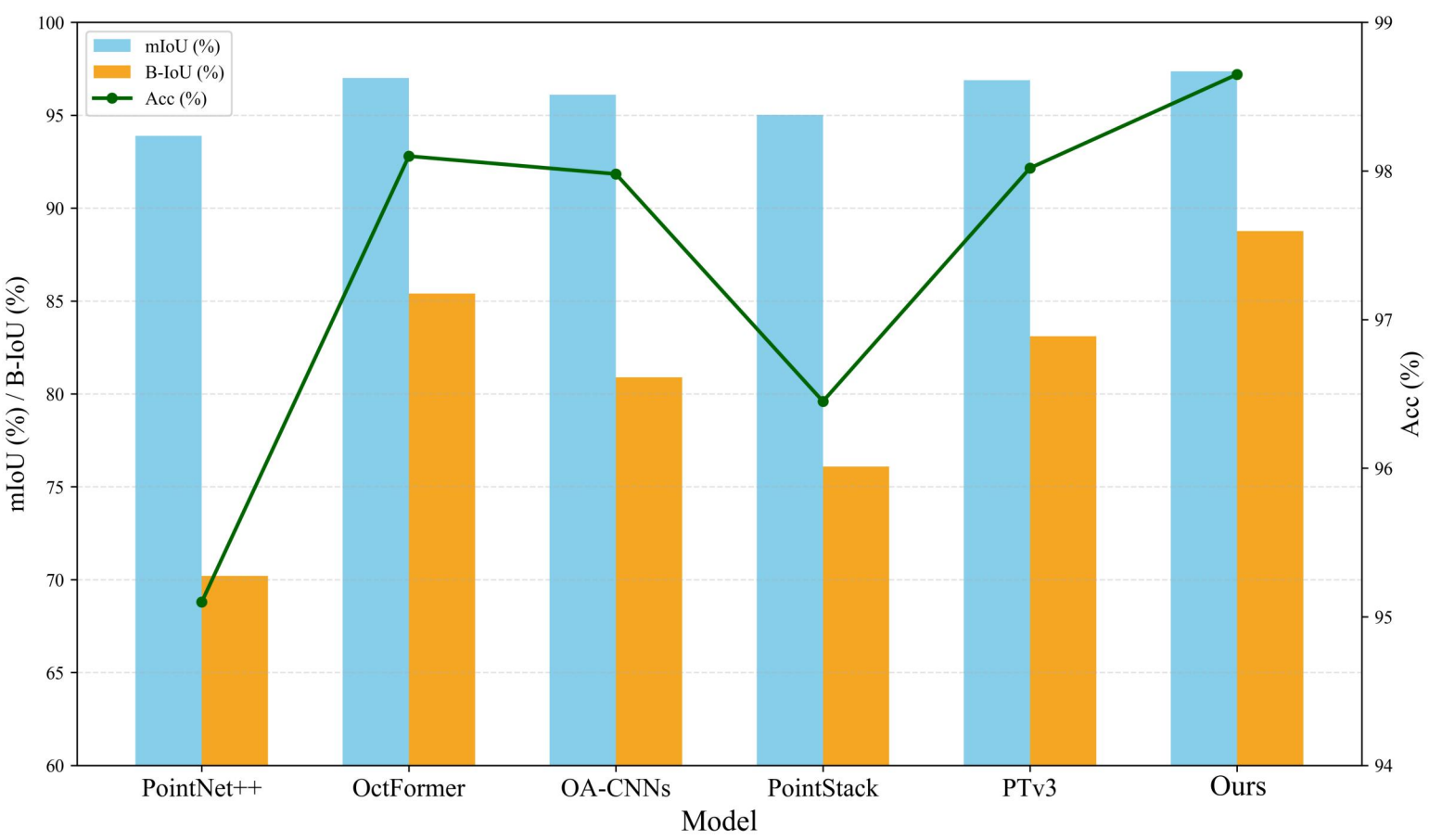

**Figure 8.** Comparative performance of different point cloud segmentation models. Acc, accuracy; mIoU, mean Intersection over Union; B-IoU, boundary Intersection over Union. Higher values indicate better segmentation performance.

## 3.3 Visual Analysis of Segmentation Results

To further analyze segmentation differences among models in complex pigsty scenarios, this study selects representative pig body point cloud samples from the test set for visual comparison, as shown in Figure 9. Overall, all models can identify the main body of the pig; however, significant differences exist in segmentation quality at contact regions between the pig body and background structures such as railings and the floor. PointNet++ and PointStack are prone to misclassifying background points as pig body points, erroneously deleting pig body edges, or producing discontinuous local boundaries in areas where the pig's head is close to a railing, the side of the torso overlaps with a fence, or the legs are near the floor. These errors manifest as boundary adhesion, contour loss, and incorrect category attribution. Compared with PointNet++, OA-CNNs and PTv3 show some improvement in boundary regions; nevertheless, when there are large variations in point cloud density across adjacent areas or when the pig body is close to background structures, local mis-segmentation still occurs.

In contrast, the proposed method produces clearer pig body contours and less background residue in these challenging areas. In regions where the pig's head contacts a railing, the method more effectively distinguishes slender railing points from pig body points, reducing misclassification of railing points as part of the pig body. In areas where the legs meet the floor, the model better preserves limb edges and reduces mis-segmentation caused by adhesion of floor points. In regions where the side of the torso overlaps with a fence, boundary continuity is improved, and the outer contour of the pig body is more complete. These visual results align with the upward trend of the *B-IoU* metric in Table 3, indicating that the advantages of the proposed method are primarily in complex contact boundary areas rather than overall classification of the main body.

This phenomenon can be attributed to the introduction of soft distance boundary pseudo-labels and the bidirectional cross-boundary semantic module on the basis of multi-scale feature representation. The former mitigates training instability caused by abrupt changes in hard labels near boundaries through continuous boundary supervision, whereas the latter facilitates information interaction between boundary features and semantic features, enabling the model to extract more discriminative contact-interface features. Consequently, under conditions of uneven point cloud density, local occlusion, and background adhesion, the proposed method produces more stable boundary segmentation results.

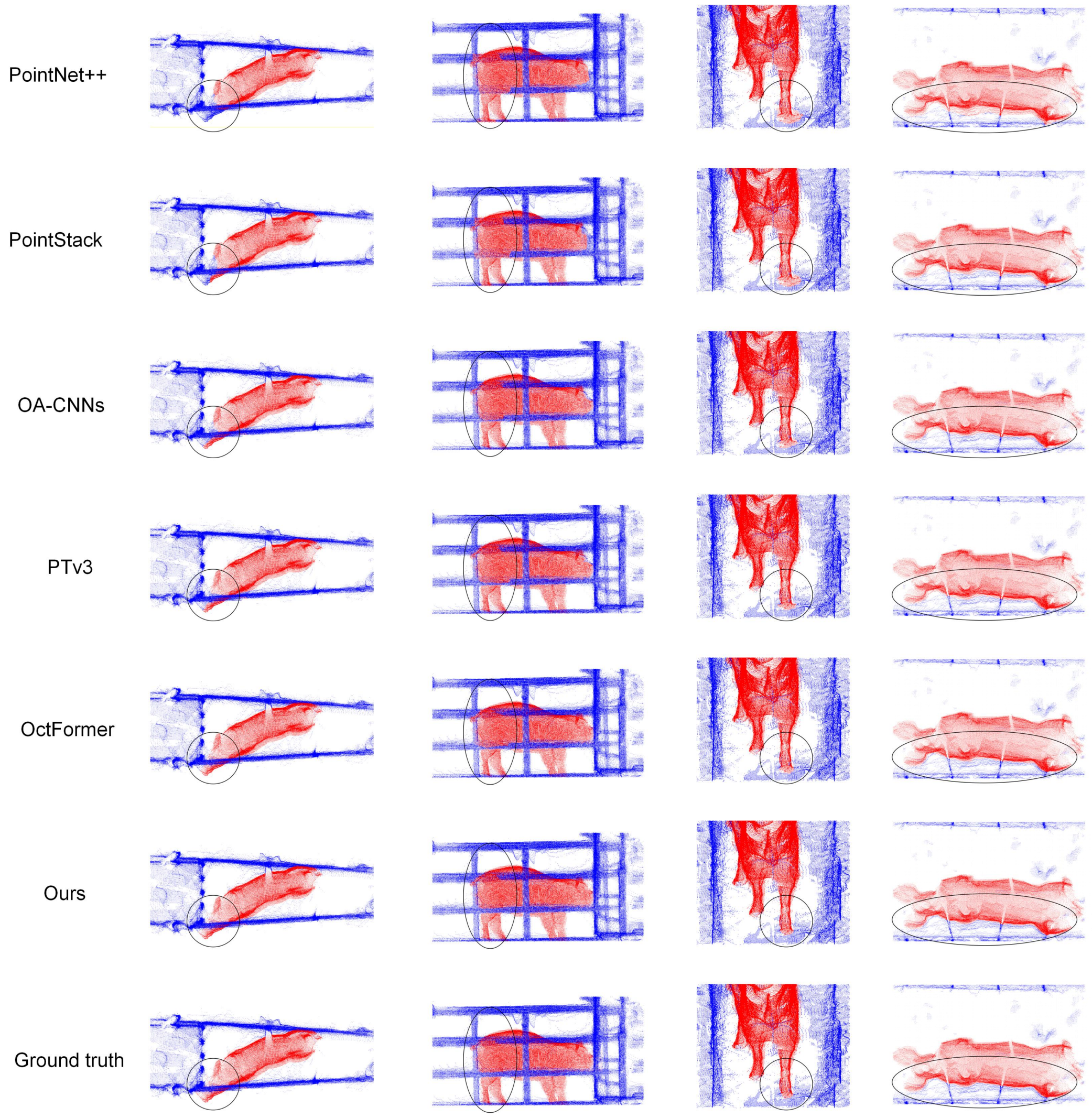

**Figure 9.** Qualitative comparison of segmentation results obtained by different point cloud segmentation models. The comparison illustrates segmentation differences in boundary-contact regions between the pig body and background structures, including railings, fences, and the floor.

## 3.4 Ablation Study on Network Modules

To evaluate the contributions of each key module to model performance, this study conducts a module ablation experiment. The results are summarized in Table 4. Model A serves as the baseline, using OctFormer as the backbone and incorporating the original boundary semantic module. In this baseline structure, the boundary branch primarily functions as a query to guide semantic attention computation, and boundary information is not fully integrated into the semantic feature update. Model B adds the Bidirectional Cross-Boundary Attention (BCBA) module to the baseline, enhancing bidirectional information interaction between semantic and boundary features. Model C incorporates the Soft Distance Boundary Pseudo-Label (SDBPL) into the baseline, providing continuous boundary supervision. The final model combines both BCBA and SDBPL.

**Table 4.** Ablation study results for network modules.

| Model | Baseline | BCBA | SDBPL | Acc (%)↑ | mIoU (%)↑ | B-IoU (%)↑ |
|---|---|---|---|---|---|---|
| A | √ | | | 98.15 | 97.06 | 85.98 |
| B | √ | √ | | 98.53 | 97.28 | 88.02 |
| C | √ | | √ | 98.22 | 97.13 | 86.98 |
| Proposed | √ | √ | √ | **98.65** | **97.36** | **88.77** |

**Note:** BCBA, Bidirectional Cross-Boundary Attention; SDBPL, Soft Distance Boundary Pseudo-Label; Acc, accuracy; mIoU, mean Intersection over Union; B-IoU, boundary Intersection over Union. √ indicates that the corresponding module was included. ↑ indicates that a higher value represents better performance.

As shown in Table 4, the baseline model (Model A) already achieves high overall segmentation performance, with *Acc* and *mIoU* reaching 98.15% and 97.06%, respectively. However, its *B-IoU* is 85.98%, indicating that relying solely on the OctFormer backbone and the unidirectional boundary guidance mechanism remains insufficient to fully resolve boundary confusion in complex contact regions. These results further suggest that, even when overall semantic segmentation accuracy is high, errors in boundary regions continue to be the primary factor limiting further improvement of the model.

After incorporating BCBA into the baseline model, Model B achieves *Acc*, *mIoU*, and *B-IoU* values of 98.53%, 97.28%, and 88.02%, respectively. Notably, its *B-IoU* is 2.04 percentage points higher than that of Model A, representing a greater improvement compared with *Acc* and *mIoU*. This result demonstrates that bidirectional interaction between semantic and boundary features effectively enhances the model's ability to represent contact boundaries. Unlike unidirectional boundary guidance, BCBA allows both boundary and semantic features to participate in information updates, facilitating the transfer of fine-grained structural information at junctions between the pig body, railings, and the floor to the semantic branch, thereby improving category discrimination in boundary regions.

After adding SDBPL, Model C achieves *Acc*, *mIoU*, and *B-IoU* values of 98.22%, 97.13%, and 86.98%, respectively, all showing improvements over Model A. In particular, *B-IoU* increases by 1.00 percentage point, indicating that soft distance boundary supervision mitigates learning instability caused by hard boundary labels in category transition areas, providing a continuous supervision signal that gradually decays from the boundary to non-boundary regions. However, compared with BCBA, the performance gain from SDBPL alone is relatively modest, suggesting that although boundary softening at the supervision level improves boundary prediction, semantic-boundary interaction at the feature level remains more critical for discriminating features in areas where the pig body closely contacts the background.

When BCBA and SDBPL are applied simultaneously, the proposed method achieves optimal results, with *Acc*, *mIoU*, and *B-IoU* reaching 98.65%, 97.36%, and 88.77%, respectively. Compared with the baseline model (Model A), these three metrics improve by 0.50, 0.30, and 2.79 percentage points, respectively. Relative to models using only BCBA or SDBPL, the proposed method demonstrates superior overall performance. These results indicate that BCBA and SDBPL exert complementary effects: BCBA enhances interaction between boundary and semantic information at the network structure level, while SDBPL mitigates training uncertainty caused by boundary blurring and point cloud density variations at the supervision signal level. The synergy between these modules enables the model to achieve stronger boundary localization and segmentation robustness in complex pigsty environments.

## 3.5 Ablation Study on Loss Functions

To further analyze the impact of loss function design on model performance, this study conducts ablation experiments on loss functions using the network with both BCBA and SDBPL incorporated. For the semantic branch, either cross-entropy (CE) loss or a combined CE + Dice loss is used. For the boundary branch, either binary cross-entropy (BCE) loss or a combined BCE + Dice loss is used. The experimental results are summarized in Table 5.

**Table 5.** Ablation study results for loss functions.

| No. | Semantic branch | Boundary branch | Acc (%)↑ | mIoU (%)↑ | B-IoU (%)↑ |
|---|---|---|---|---|---|
| 1 | CE | BCE | 98.48 | 97.24 | 87.69 |
| 2 | CE | BCE + Dice | 98.53 | 97.29 | 88.33 |

| 3 | CE + Dice | BCE | 98.54 | 97.27 | 87.98 |
|---|---|---|---|---|---|
| 4 | CE + Dice | BCE + Dice | **98.65** | **97.36** | **88.77** |

**Note:** CE, cross-entropy; BCE, binary cross-entropy; Dice, Dice loss; Acc, accuracy; mIoU, mean Intersection over Union; B-IoU, boundary Intersection over Union. ↑ indicates that a higher value represents better performance.

In Experiment 1, only CE and BCE losses were used as the baseline configuration, resulting in *Acc*, *mIoU*, and *B-IoU* values of 98.48%, 97.24%, and 87.69%, respectively. In Experiment 2, after introducing Dice loss to the boundary branch, *Acc*, *mIoU*, and *B-IoU* increased to 98.53%, 97.29%, and 88.33%, representing gains of 0.05, 0.05, and 0.64 percentage points compared with Experiment 1. Among these metrics, the improvement in *B-IoU* was the most pronounced, indicating that Dice loss strengthens the regional overlap constraint between boundary predictions and soft boundary pseudo-labels. Because boundary points constitute a relatively small proportion of the overall point cloud, relying solely on BCE is susceptible to class imbalance. Dice loss optimizes the prediction region from the perspective of set overlap, providing stronger constraints on minority boundary regions and thereby improving boundary segmentation performance.

Experiment 3 adopted the CE + Dice combined loss in the semantic branch, while the boundary branch still used BCE. Compared with Experiment 1, *Acc*, *mIoU*, and *B-IoU* increased by 0.06, 0.03, and 0.29 percentage points respectively, indicating that introducing the Dice loss in the semantic branch helps improve the overall consistency between the predicted and true labels of the main region, but the improvement of the boundary metrics is less than that when the Dice loss is introduced in the boundary branch. This result shows that for the pig point cloud segmentation task with contact boundaries as the main source of error, the regional overlap optimization of the boundary branch is more critical for *B-IoU*.

When the combined CE + Dice loss is applied to both the semantic and boundary branches, Experiment 4 achieves optimal results, with *Acc*, *mIoU*, and *B-IoU* reaching 98.65%, 97.36%, and 88.77%, respectively. Compared with Experiment 1, these three metrics increase by 0.17, 0.12, and 1.08 percentage points, respectively. Relative to Experiments 2 and 3, Experiment 4 also demonstrates further improvements across all metrics. These results indicate that CE/BCE provides stable point-wise classification supervision, while Dice loss further enforces overall overlap consistency between predicted and ground-truth regions. By combining these losses, the model simultaneously improves segmentation quality in the pig's main body region and enhances recognition of complex boundary regions.

Based on the above experimental results, it is evident that the performance improvement of the proposed method arises not only from octree-based multi-scale feature extraction but also from boundary supervision, semantic-boundary interaction, and loss function design. Specifically, SDBPL provides continuous supervision signals for contact boundaries, BCBA facilitates bidirectional interaction between boundary and semantic features, and Dice loss further strengthens overlapping optimization for minority boundary regions. The combined effect of these three components enables the model to produce more stable segmentation results in boundary regions where mis-segmentation is most likely to occur in real pigsty environments.

# 4 Discussion

The results of this study indicate that the pig point cloud segmentation method based on boundary feature analysis improves segmentation between pig body point clouds and background structures in complex pigsty environments. Compared with comparative models such as PointNet++, OctFormer, OA-CNNs, PointStack, and PTv3, the proposed method achieves the best performance across the three metrics of *Acc*, *mIoU*, and *B-IoU*. Among these metrics, the improvements in *Acc* and *mIoU* are relatively limited, whereas the improvement in *B-IoU* is more pronounced. This result suggests that when the main pig body region can be accurately recognized, differences in model performance are mainly concentrated at contact boundaries between the pig body and background structures such as the floor, fences, and railings. For real pigsty point cloud segmentation tasks, *B-IoU* better reflects the practical value of the model in challenging regions than overall accuracy, because areas where the pig's head is close to railings, the legs are close to the floor, or the side of the torso overlaps with fences are major sources of error propagation in subsequent point cloud completion and body size measurement.

From a structural perspective, the effectiveness of the proposed method primarily stems from the hierarchical organization capability of the octree Transformer for sparse 3D point clouds. Pig point clouds are characterized by disorder, sparsity, and uneven local density, with pig bodies, railings, the floor, and noise points often interlaced in space. When relying solely on feature aggregation of unstructured point sets, the model is prone to misclassification in regions with local density variations or background adhesion. The octree structure organizes point clouds into non-empty nodes with spatial hierarchical relationships and performs attention computations within local windows, reducing computational burden while preserving local geometric relationships. Further, multi-scale feature fusion allows the model to simultaneously leverage high-level semantic information and low-level boundary details, avoiding problems such as overly thick boundaries or local structure fragmentation caused by using single-scale features alone.

The key difference between the proposed method and general point cloud semantic segmentation networks lies in its explicit modeling of contact boundaries. Most existing studies on pig body point cloud segmentation improve overall feature extraction by adjusting the sampling radius, increasing network depth, or modifying pooling strategies. Although these methods can improve segmentation performance in main body regions, their ability to model boundary uncertainty at contact regions between the pig body and railings or the floor remains limited. In contrast, the proposed method treats boundary regions as the primary optimization target. Specifically, the soft distance boundary pseudo-label converts discrete boundary supervision into continuous boundary constraints, enabling the model to learn the distance relationship between points and boundaries rather than simply determining whether a point belongs to a boundary. Meanwhile, the bidirectional cross-boundary semantic module facilitates information exchange between boundary and semantic features, allowing boundary structural information to participate in semantic feature updates. By combining these two components, the model can not only capture local details but also establish a clearer basis for category discrimination in pig body–background contact regions.

The ablation experiments further support this interpretation. When BCBA is introduced alone, the improvement in *B-IoU* is greater than those in *Acc* and *mIoU*, indicating that bidirectional interaction between semantic and boundary features primarily improves boundary regions rather than main body regions. Introducing SDBPL alone also improves performance, suggesting that continuous boundary supervision mitigates training instability caused by hard-boundary labels in category transition regions. The best performance is achieved when both modules are used simultaneously, indicating that boundary supervision and boundary–semantic feature interaction exert complementary effects. The loss function ablation results also show that the model performs best when Dice loss is introduced into both the semantic and boundary branches, which is consistent with the task characteristics of a low proportion of boundary points and an imbalanced class distribution. Dice loss constrains prediction results from the perspective of regional overlap, thereby increasing the optimization weight of minority boundary regions and improving the integrity of boundary prediction.

From an application perspective, high-quality pig body point cloud segmentation is a prerequisite for subsequent point cloud completion, keypoint detection, and automatic body size measurement. If background points such as railings and the floor are incorrectly retained as pig body points during front-end segmentation, the subsequent completion model may learn incorrect local structures, leading to background residues or abnormal protrusions in the completed point clouds. Conversely, if the edges of the pig's legs, abdomen, or back are erroneously removed, the measurement stability of parameters such as chest circumference, abdominal circumference, and hip circumference may be affected, as these measurements rely on the integrity of cross-sectional point clouds. Therefore, the improvement in boundary regions achieved by the proposed method is significant not only in terms of segmentation metrics but also in reducing the risk of error propagation during subsequent 3D phenotypic measurement. In large-scale pig farms, pig movement, occlusion, and contact with railings are difficult to avoid completely. A segmentation method capable of stably handling boundary adhesion is therefore more suitable as the front-end module of an automated phenotypic measurement system.

It should be noted that the generalization ability and engineering applicability of the proposed method still require further validation. First, the experimental data were primarily collected from specific pig farms under fixed camera deployment conditions, where railing morphology, passage width, camera height, and point cloud density distributions were relatively stable. Changes in the collection environment, such as using different pig farms, adjusting the number of cameras, or adopting alternative pen structures, may reduce the model's adaptability to background morphology and boundary distributions. Second, the soft-distance boundary pseudo-labels depend on neighborhood radius and distance attenuation parameters. Significant changes in point cloud density could cause these fixed parameters to produce overly wide or narrow boundary supervision ranges, potentially affecting model learning. Third, manual annotation at contact points between pigs, railings, and the floor inevitably involves some subjectivity. Labeling errors in boundary-blurred regions may further influence the construction of soft-boundary pseudo-labels. Future research should conduct generalization tests across multiple pig farms, devices, and viewpoint configurations, and further analyze the effects of boundary radius, attenuation scale, and input point cloud density on model performance.

In addition, the current validation of the proposed method primarily focuses on segmentation accuracy, with analysis of model complexity, inference speed, and deployment cost remaining insufficient. Although the octree Transformer and boundary semantic enhancement module improve boundary segmentation quality, they may incur higher computational overhead compared with lightweight networks. In practical aquaculture scenarios, the segmentation model must operate in coordination with modules such as multi-camera acquisition, point cloud registration, point cloud completion, and body size measurement. Therefore, real-time performance and hardware cost are also critical factors for practical deployment. In the future, while maintaining boundary segmentation accuracy, lightweight boundary enhancement modules, dynamic window attention, or knowledge distillation strategies could be explored to improve the feasibility of this method for large-scale on-site deployment in pig farms.

# 5 Conclusions

To address challenges in real pigsty environments, such as close contact, blurred boundaries, and local adhesion between pig point clouds and background structures like the floor, fences, and railings, this study proposes a pig point cloud segmentation method based on boundary feature analysis. The method employs the Octree Transformer as the backbone network, constructing a point cloud feature representation that integrates local geometric details and global semantic context through octree convolution, self-attention encoding, and multi-scale feature fusion. Building on this, a soft-distance boundary pseudo-label and a bidirectional cross-boundary semantic module are introduced to enhance the model's discriminative capability in pig–background contact areas through two mechanisms: continuous boundary supervision and semantic–boundary feature interaction.

The experimental results show that the proposed method achieves an *Acc* of 98.65%, an *mIoU* of 97.36%, and a *B-IoU* of 88.77% on the pig point cloud segmentation test set. Its overall performance is superior to that of comparative models, including PointNet++, OctFormer, OA-CNNs, PointStack, and PTv3. Compared with the second-best model, OctFormer, the proposed method improves B-IoU by 3.37 percentage points, indicating that its advantage is primarily reflected in the fine-grained segmentation of complex contact boundary regions between the pig body and background structures such as railings and the floor. The visualization results further show that the proposed method reduces background residues and boundary adhesion in regions where the pig's head contacts railings, the legs meet the floor, and the side of the torso overlaps with fences. The ablation experiments further verify the effectiveness of BCBA and SDBPL. The model achieves optimal performance when both modules are used together, indicating that the semantic–boundary bidirectional interaction mechanism and soft distance boundary supervision are complementary.

In summary, the proposed method effectively mitigates background adhesion and boundary mis-segmentation in complex pigsty point clouds, providing more reliable pig body point cloud inputs for subsequent tasks such as point cloud completion, keypoint detection, and automatic body size measurement. It should be noted, however, that the experiments were primarily conducted under specific pig farms and fixed collection conditions. The model's generalization across varying pigsty structures, railing shapes, camera layouts, and point cloud densities requires further validation. Future research will focus on conducting robustness tests across multiple pig farms, devices, and viewpoint configurations, as well as combining lightweight network design with boundary parameter sensitivity analysis to enhance the feasibility of deploying this method in large-scale pig farming operations.

## Supplementary Materials

No supplementary material files are provided with this manuscript.

## Ethics Approval Statement

This study involved non-invasive observational point cloud data collection from pigs during routine commercial farm management. No invasive intervention, experimental treatment, or additional animal handling was performed. Data collection was conducted with permission from the participating farms, and all procedures followed routine husbandry practices and relevant animal welfare principles.

## Author Contributions

Conceptualization, Z.X., F.S., Q.L. and W.M.; methodology, Z.X., F.S. and W.M.; software, Z.X., F.S. and X.Q.; validation, Z.X., F.S., Z.W. and M.G.; formal analysis, Z.X., F.S. and X.Q.; investigation, Z.X., Z.W., M.G. and Y.F.; resources, Q.L. and W.M.; data curation, Z.X., F.S., Z.W., M.G. and Y.F.; writing—original draft preparation, Z.X. and F.S.; writing—review and editing, X.Q., Z.W., M.G., Y.F., S.X.Y., Q.L. and W.M.; visualization, Z.X., F.S. and X.Q.; supervision, S.X.Y., Q.L. and W.M.; project administration, Q.L. and W.M.; funding acquisition, Q.L. and W.M. Z.X. and F.S. contributed equally to this work. All authors have read and agreed to the published version of the manuscript.

## Funding

This work was supported by Sichuan Science and Technology Program (No2021ZDZX0011) and the Special Project for Outstanding Scientist Cultivation of Beijing Academy of Agriculture and Forestry Sciences(JKZX202214)

## Data Availability and Statistical Reporting

The raw point cloud datasets generated and analyzed during the current study are not publicly available because they contain farm-specific acquisition information and individual animal identification records, and their release is restricted by data-use and farm-management agreements. Anonymized example point clouds and additional data may be made available from the corresponding author upon reasonable request and with permission from the participating farms.

The source code, model configuration files, training and evaluation scripts, implementation details, and trained model parameters are not publicly available at this stage because they are part of an ongoing research project and are subject to institutional data-management and project-use restrictions. Relevant implementation details and trained model parameters may be made available from the corresponding author upon reasonable request for academic and non-commercial research purposes, subject to applicable restrictions.

All experiments were conducted using the training, validation, and test splits described in the Materials and Methods section. The dataset was divided at the individual-pig level to reduce the risk of data leakage across subsets. Samples with severe posture distortion, motion blur, or sensor abnormalities were excluded before dataset partitioning according to the predefined quality-screening procedure. No data imputation was performed. Model performance was evaluated using accuracy, mean Intersection over Union, and boundary Intersection over Union on the independent test set. No formal null-hypothesis testing was conducted, and therefore no multiple-testing correction was applied. Statistical summaries and evaluation metrics were calculated using Python, and model training and testing were implemented using the software environment described in the Experimental Setup section.

## Conflict of Interest Disclosure

The authors declare that they have no known competing financial interests or personal relationships that could have appeared to influence the work reported in this manuscript. All authors have reviewed the journal's conflict of interest policy and confirm that there are no relevant commercial or other relationships to disclose.